\documentclass[runningheads]{llncs}
\usepackage{times}
\usepackage{epsfig}
\usepackage{graphicx}
\usepackage{amsmath}
\usepackage{amssymb}
\usepackage{amssymb}
\usepackage{amsfonts}
\usepackage{cite}
\usepackage{microtype} 
\usepackage{caption} 

\usepackage{tabularx}
\usepackage{booktabs,calc}
\usepackage{multirow}
\usepackage{hhline}
\usepackage{array}
\usepackage{float}
\usepackage[caption=false]{subfig}
\restylefloat{table}
\usepackage{makecell}
\usepackage{color}
\usepackage{mathtools}
\usepackage{hyperref}

\hypersetup{
  colorlinks   = true, 
  urlcolor     = blue, 
  linkcolor    = red, 
  citecolor   = green 
}

\usepackage[width=122mm,left=12mm,paperwidth=146mm,height=193mm,top=12mm,paperheight=217mm]{geometry}

\newcolumntype{L}[1]{>{\raggedright\let\newline\\\arraybackslash\hspace{0pt}}m{#1}}
\newcolumntype{C}[1]{>{\centering\let\newline\\\arraybackslash\hspace{0pt}}m{#1}}
\newcolumntype{R}[1]{>{\raggedleft\let\newline\\\arraybackslash\hspace{0pt}}m{#1}}

\newcommand{\ie}{\mbox{\emph{i.e.}}}
\newcommand{\eg}{\mbox{\emph{e.g.}}}

\DeclareMathOperator*{\argmax}{argmax}

\begin{document}
\pagestyle{headings}
\mainmatter
\def\ECCV16SubNumber{1}  

\title{Multi-Person Pose Estimation with Local Joint-to-Person Associations} 

\titlerunning{Multi-Person Pose Estimation with Local Joint-to-Person Associations}

\authorrunning{Umar Iqbal and Juergen Gall}

\author{Umar Iqbal and Juergen Gall}


\institute{Computer Vision Group\\
	University of Bonn, Germany\\
	\email{ \{uiqbal,gall\}@iai.uni-bonn.de}
}

\maketitle

\begin{abstract}
Despite of the recent success of neural networks for human pose estimation, current approaches are limited to pose estimation of a single person and cannot handle humans in groups or crowds. In this work, we propose a method that estimates the poses of multiple persons in an image in which a person can be occluded by another person or might be truncated. To this end, we consider multi-person pose estimation as a joint-to-person association problem. We construct a fully connected graph from a set of detected joint candidates in an image and resolve the joint-to-person association and outlier detection  using integer linear programming. Since solving joint-to-person association jointly for all persons in an image is an NP-hard problem and even approximations are expensive, we solve the problem locally for each person. On the challenging MPII Human Pose Dataset for multiple persons, our approach achieves the accuracy of a state-of-the-art method, but it is 6,000 to 19,000 times faster.   
\end{abstract}

\section{Introduction}
\begin{figure*}[t!]
\centering
\captionsetup[figure]{skip=0pt}
\includegraphics[scale=1]{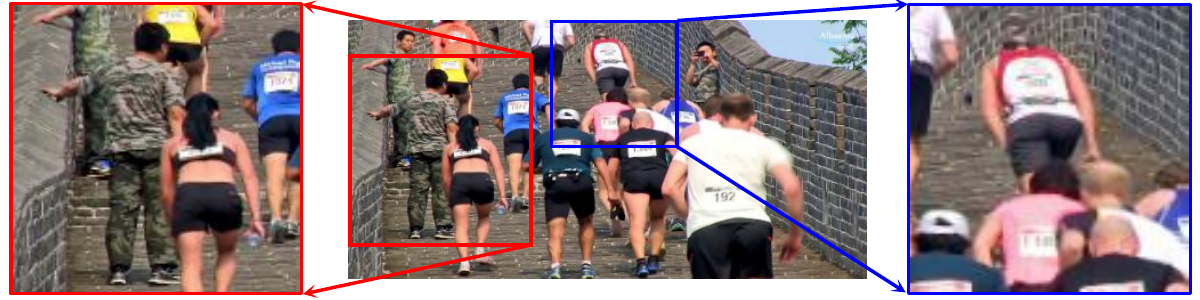}
\caption{Example image from the multi-person subset of the MPII Pose Dataset~\cite{andriluka_cvpr2014}.}
\label{fig:motivation_image}
\end{figure*}

Single person pose estimation has made a remarkable progress over the past few years. This is mainly due to the availability of deep learning based methods for detecting joints \cite{carreira2015human, pishchulin2015deepcut, wei2016convolutional, tompson_cvpr2015, insafutdinov2016deepercut}. While earlier approaches in this direction \cite{chen_nips2014, tompson2014joint, tompson_cvpr2015} combine the body part detectors with tree structured graphical models, more recent methods \cite{carreira2015human, pishchulin2015deepcut, wei2016convolutional,newell2016eccv, bulat2016human, rafi2016bmvc} demonstrate that spatial relations between joints can be directly learned by a neural network without the need of an additional graphical model. These approaches, however, assume that only a single person is visible in the image and the location of the person is known a-priori. Moreover, the number of parts are defined by the network, \eg, full body or upper body, and cannot be changed. For realistic scenarios such assumptions are too strong and the methods cannot be applied to images that contain a number of overlapping and truncated persons. An example of such a scenario is shown in Figure~\ref{fig:motivation_image}. 

In comparison to single person human pose estimation benchmarks, multi-person pose estimation introduces new challenges. The number of persons in an image is unknown and needs to be correctly estimated, the persons occlude each other and might be truncated, and the joints need to be associated to the correct person. The simplest approach to tackle this problem is to first use a person detector and then estimate the pose for each detection independently~\cite{pishchulin2012articulated, gkioxari2014using, chen2015parsing}. This, however, does not resolve the joint association problem of two persons next to each other or truncations. Other approaches estimate the pose of all detected persons jointly~\cite{eichner2010we, Ladicky_2013_CVPR}. In~\cite{pishchulin2015deepcut} a person detector is not required. Instead body part proposals are generated and connected in a large graph. The approach then solves the labeling problem, the joint-to-person association problem and non-maximum suppression jointly. While the model proposed in~\cite{pishchulin2015deepcut} can be solved by integer linear programming and achieves state-of-the-art results on a very small subset of the MPII Human Pose Dataset, the complexity makes it infeasible for a practical application. As reported in \cite{insafutdinov2016deepercut}, the processing of a single image takes about 72 hours. 

In this work, we address the joint-to-person association problem using a densely connected graphical model as in~\cite{pishchulin2015deepcut}, but propose to solve it only locally. To this end, we first use a person detector and crop image regions as illustrated in  Figure~\ref{fig:motivation_image}. Each of the regions contains sufficient context, but only the joints of persons that are very close. We then solve the joint-to-person association for the person in the center of each region by integer linear programming (ILP). The labeling of the joints and non-maxima suppression are directly performed by a convolutional neural network. We evaluate our approach on the MPII Human Pose Dataset for multiple persons where we slightly improve the accuracy of \cite{pishchulin2015deepcut} while reducing the runtime by a factor between 6,000 and 19,000.

\section{Related Work}
Human pose estimation has generally been addressed under the assumption that only a single person is visible. For this, earlier approaches formulate the problem in a graphical model where interactions between body parts are modelled in a tree structure combined with local observations obtained from discriminatively trained part detectors \cite{Felzenszwalb_ijcv2005, tran2010improved, andriluka_ijcv2012, pishchulin_cvpr2013, Wang_CVPR2013, yang_tpami2014, dantone_tpami2014}. While the tree-structured models provide efficient inference, they struggle to model long-range characteristics of the human body. With the progress in convolutional neural network architectures, more recent works adopt CNNs to obtain stronger part detectors but still use graphical models to obtain coherent pose estimates \cite{chen_nips2014, tompson2014joint, tompson_cvpr2015}. 

The state-of-the-art approaches, however, demonstrate that graphical models are of little importance in the presence of strong part detectors since the long-range relationships of the body parts can be directly incorporated in the part detectors \cite{pishchulin2015deepcut, carreira2015human, wei2016convolutional, insafutdinov2016deepercut, newell2016eccv, bulat2016human, rafi2016bmvc}. In \cite{wei2016convolutional, newell2016eccv, bulat2016human} multi-staged CNN architectures are proposed where each stage of the network takes as input the score maps of all parts from its preceding stage. This provides additional information about the interdependence, co-occurrence, and context of parts to each stage, and thereby allows the network to implicitly learn image dependent spatial relationships between parts. Similarly, instead of a multi-staged architecture, \cite{insafutdinov2016deepercut} proposes to use a very deep network that inherently results in large receptive fields and therefore allows to use contextual information around the parts. All of these methods report impressive results for single person pose estimation without an additional graphical model for refinement. 

In contrast to the single person pose estimation, multi-person pose estimation poses a significantly more complex problem, and only a few works have focused in this direction \cite{eichner2010we, sun2011articulated, pishchulin2012articulated, yang_tpami2014, ladicky2013human, gkioxari2014using, chen2015parsing, belagiannis2015tpami, pishchulin2015deepcut, insafutdinov2016deepercut}. \cite{yang_tpami2014} and \cite{sun2011articulated} perform non-maximum suppression on the marginals obtained using a graphical model to generate multiple pose hypotheses in an image. The approaches, however, can only work in scenarios where persons are significantly distant from each other and consider only the fully visible persons. The methods in \cite{pishchulin2012articulated, gkioxari2014using, chen2015parsing} first detect the persons in an image using a person detector and then estimate the body pose for each person independently. \cite{sun2011articulated} employs a similar approach for 3D pose estimation of multiple persons in a calibrated multi-camera scenario. The approach first obtains the number of persons using a person detector and then samples the 3D poses for each person from the marginals of a 3D pictorial structure model. For every detected person, \cite{chen2015parsing} explores a range of tree structured models each containing only a subset of upper-body parts, and selects the best model based on a cost function that penalizes a model containing occluded parts. Since the search space of the models increases exponentially with the number of body parts, the approach is very expensive for full body skeletons. \cite{eichner2010we, Ladicky_2013_CVPR} define a joint pose estimation model for all detected persons, and utilize several occlusion clues to model interactions between people. All these approaches rely on a standard pictorial structure model with tree structures and cannot incorporate dependencies beyond adjacent joints.

More recently, \cite{pishchulin2015deepcut} proposed a joint objective function to solve multi-person pose estimation. The approach does not require a separate person detector or any prior information about the number of persons. Unlike earlier works it can tackle any type of occlusion or truncation. It starts by generating a set of class independent part proposals and constructs a densely connected graph from the proposals. It uses integer linear programming to label each proposal by a certain body part and assigns them to unique individuals. The optimization problem proposed in \cite{pishchulin2015deepcut} is theoretically well founded, but is an NP-Hard problem to solve and prohibitively expensive for realistic scenarios. Therefore, it limits the number of part proposals to 100. This means that the approach can estimate the poses of at most 7 fully visible persons with 14 body parts per person. Despite the restriction, the inference takes roughly 72 hours for a single image \cite{insafutdinov2016deepercut}. In \cite{insafutdinov2016deepercut}, the authors build upon the same model and propose to use stronger part detectors and image dependent spatial models along with an incremental optimization approach that significantly reduces the optimization time of \cite{pishchulin2015deepcut}. The approach, however, is still too slow for practical applications since it requires 8 minutes per image and still limits the number of proposals to a maximum of 150. 

\section{Overview}
\begin{figure*}[t!]
  \centering
  \begin{tabular}{c c}
    \includegraphics[height=0.260\linewidth]{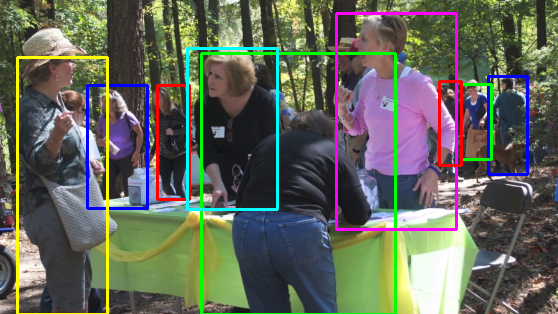} &
    \includegraphics[height=0.260\linewidth]{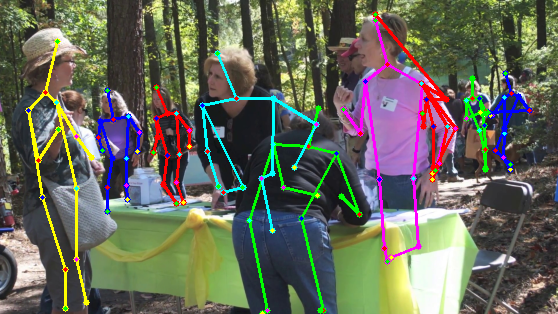} \\
    (a) & (b)
  \end{tabular}

  \begin{tabular}{c c}
    \includegraphics[height=0.260\linewidth]{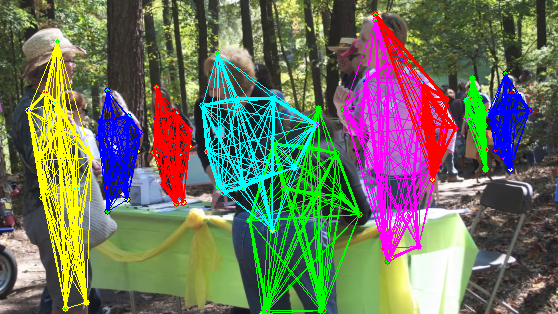} &
    \includegraphics[height=0.260\linewidth]{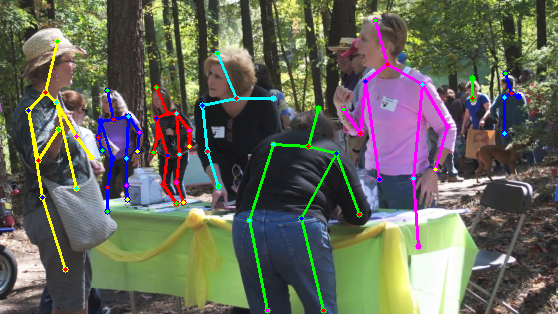} \\
    (c) & (d)
  \end{tabular}
  \caption{Overview of the proposed method. We detect persons in an image using a person detector (a). A set of joint candidates is generated for each detected person (b). The candidates build a fully connected graph (c) and the final pose estimates are obtained by integer linear programming (d). (best viewed in color)}
  \label{fig:overview}
\end{figure*}

Our method solves the problem of joint-to-person association locally for each person in the image. To this end, we first detect the persons using a person detector~\cite{ren2015faster}. For each detected person, we generate a set of joint candidates using a single person pose estimation model (Section~\ref{sec:cpm}). The candidates are prone to errors since the single person models do not take into account occlusion or truncation. In order to associate each joint to the correct person and also to remove the erroneous candidates, we perform inference locally on a fully connected graph for each person using integer linear programming (Section~\ref{sec:jtop}). Figure \ref{fig:overview} shows an overview of the proposed approach.

\section{Convolutional Pose Machines}
\label{sec:cpm}

Given a person in an image $\bold{I}$, we define its pose as a set $\mathcal{X}$ = $\{\bold{x}_j\}_{j = 1 \dots J}$ of $J=14$ body joints, where the vector $\bold{x}_j \in \mathcal{X}$ represents the 2D location $(u,v)$ of the $j^{th}$ joint in the image. The convolutional pose machines consist of a multi-staged CNN architecture with $t \in \{1 \dots T\}$ stages, where each stage is a multi-label classifier $\phi_{t}(\bold{x})$ that is trained to provide confidence maps $s^j_t \in \mathbb{R}^{w \times h}$ for each joint $j = 1 \dots J$ and the background, where $w$ and $h$ are the width and the height of the image, respectively.
 
\begin{figure*}[t!]
\centering
\captionsetup[figure]{skip=0pt}
\includegraphics[scale=1]{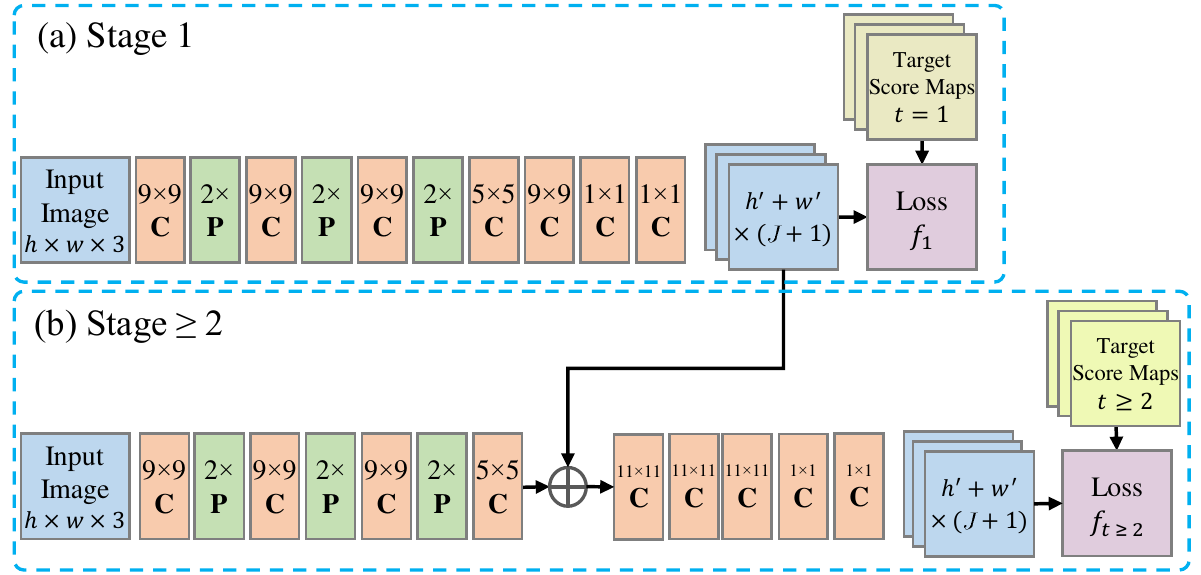}
\caption{CPM architecture proposed in \cite{wei2016convolutional}. The first stage (a) utilizes only the local image evidence whereas all subsequent stages (b) also utilize the output of preceding stages to exploit the spatial context between joints. The receptive field of stages $ t \geq 2 $ is increased by having multiple convolutional layers at the $8$ times down-sampled score maps. All stages are locally supervised and a separate loss is computed for each stage. We provide multi-person target score maps to stage 1, and single-person score maps to all subsequent stages.}
\label{fig:cpm_arch}
\end{figure*}

The first stage of the architecture uses only the local image evidence and provides the confidence scores
\begin{equation}
\phi_{t=1}(\bold{x}|\bold{I}) \rightarrow \{s^j_1(\bold{x}_j = \bold{x})\}_{j=1 \dots J+1}.
\end{equation}
Whereas, in addition to the local image evidence, all subsequent stages also utilize the contextual information from the preceding stages to produce confidence score maps
\begin{equation}
\phi_{t>1}(\bold{x|\bold{I}, \psi(\bold{x}, \bold{s}_{t-1})}) \rightarrow \{s^j_t(\bold{x}_j = \bold{x})\}_{j=1 \dots J+1},
\end{equation}
where $\bold{s}_t \in \mathbb{R}^{w \times h \times (J+1)}$ corresponds to the score maps of all body joints and the background at stage $t$, and $\psi(\bold{x}, \bold{s}_{t-1})$ indicates the mapping from the scores  $\bold{s}_{t-1}$ to the context features for location $\bold{x}$. The receptive field of the subsequent stages is increased to the extent that the context of the complete person is available. This allows to model complex long-range spatial relationships between joints, and to leverage the context around the person. The CPM architecture is completely differentiable and allows end-to-end training of all stages. Due to the multi-stage nature of CPM, the overall CNN architecture consists of many layers and is therefore prone to the problem of vanishing gradients \cite{bengio1994learning, glorot2010understanding, wei2016convolutional}. In order to solve this problem, \cite{wei2016convolutional} uses intermediate supervision by adding a loss function at each stage $t$. The CNN architecture used for each stage can be seen in Figure \ref{fig:cpm_arch}. 
In this paper we exploit the intermediate supervision of the stages during training for multi-person human pose estimation as we will discuss in the next section. 
 
\subsection{Training for Multi-Person Pose Estimation}

\begin{figure*}[t!]
\centering
\captionsetup[figure]{skip=0pt}
\includegraphics[trim={0 1.2cm 0 0},clip, scale=1]{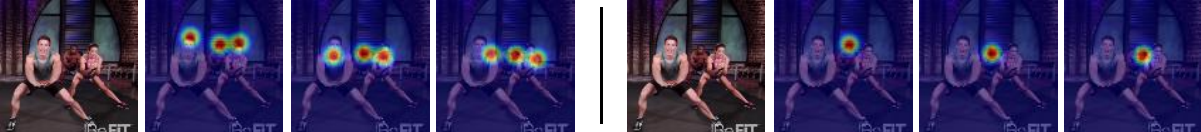}
\caption{Example of target score maps for the head, neck and left shoulder. The target score maps for the first stage include the joints of all persons (left). The target score maps for all subsequent stages only include the joints of the primary person.}
\label{fig:ideal_target_maps_example}
\end{figure*}

Each stage of the CPM is trained to produce confidence score maps for all body joints, and the loss function at the end of every stage computes the $l_2$ distance between the predicted confidence scores and the target score maps. The target score maps are modeled as Gaussian distributions centered at the ground-truth locations of the joints. 
For multi-person pose estimation, the aim of the training is to focus only on the body joints of the detected person, while suppressing joints of all other overlapping persons. We do this by creating two types of target score maps. For the first stage, we model the target score maps by a sum of Gaussian distributions for the body joints of all persons appearing in the bounding box enclosing the primary person that appears roughly in the center of the bounding box. For the subsequent stages, we model only the joints of the primary person. 
An example of target score maps for different stages can be seen in Figure \ref{fig:ideal_target_maps_example}. 

\begin{figure*}[t!]
\centering
\captionsetup[figure]{skip=0pt}
\includegraphics[trim={0 3.75cm 0 0},clip, scale=0.95]{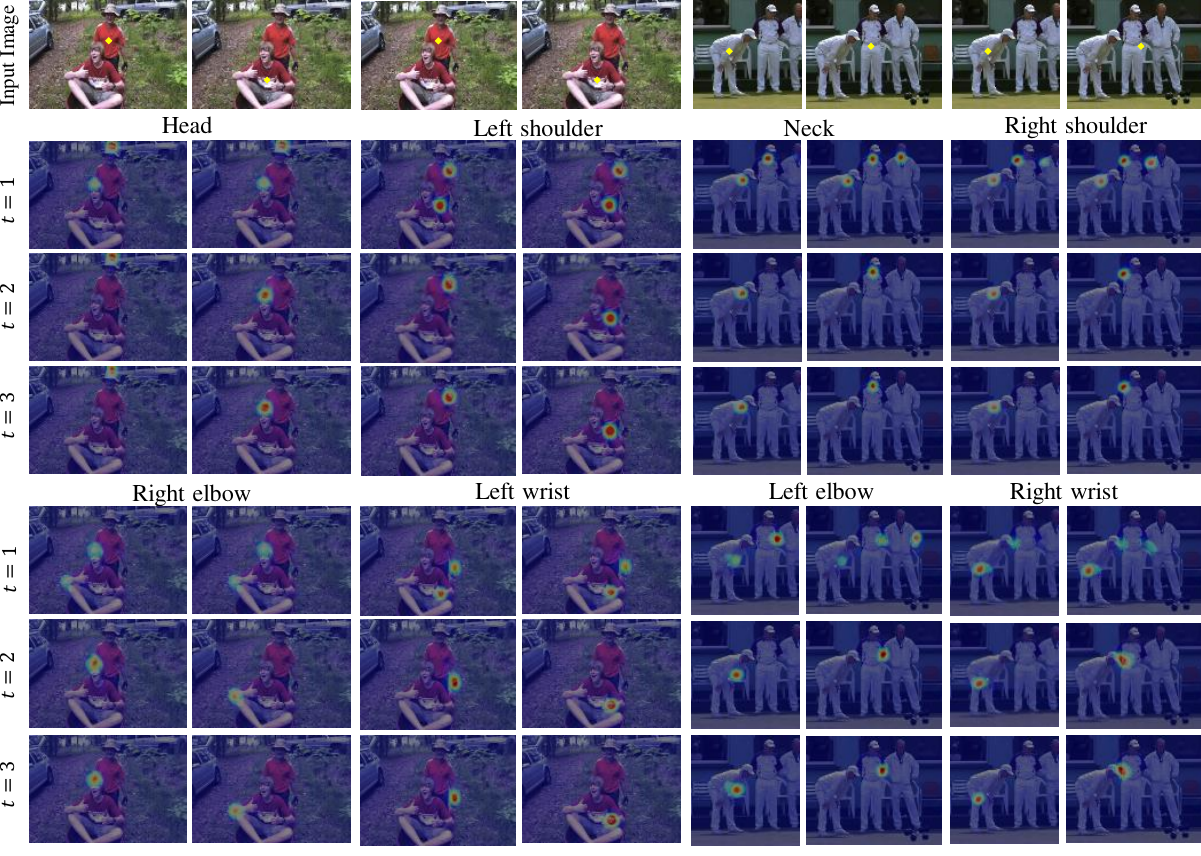}
\caption{Examples of score maps provided by different stages of the CPM. The first stage of CPM uses only local image evidence and therefore provides high confidence scores for the joints of all persons in the image. Whereas all subsequent stages are trained to provide high confidence scores only for the joints of the primary person while suppressing the joints of other persons. The primary person is highlighted by a yellow dot in the first row. (best viewed in color)}
\label{fig:example_score_maps}
\end{figure*}

Figure \ref{fig:example_score_maps} shows some examples how the inferred score maps evolve as the number of stages increases. In \cite{wei2016convolutional}, the pose of the person is obtained by taking the maximum of the inferred score maps, \ie, $\bold{x}_{j} = \argmax_{\bold{x}} s^j_{T}(\bold{x})$.

This, however, assumes that all joints are visible in the image and results in erroneous estimates for invisible joints and can wrongly associate joints of other nearby persons to the primary person. Instead of taking the maximum, we sample $N$ candidates from each inferred score map $s^j_{T}$ and resolve the joint-to-person association and outlier removal by integer linear programming.   

\section{Joint-to-Person Association}\label{sec:jtop}
We solve the joint-to-person association using a densely connected graphical model as in \cite{pishchulin2015deepcut}. The model proposed in \cite{pishchulin2015deepcut}, however, aims to resolve joint-to-person associations together with proposal labeling globally for all persons, which makes it very expensive to solve. In contrast, we propose to solve this problem locally for each person. We first briefly summarize the DeepCut method~\cite{pishchulin2015deepcut} in Section \ref{sec:deepcut}, and then describe the proposed local joint-to-person association model in Section \ref{sec:local_joint_to_person_association}. 

\subsection{DeepCut}
\label{sec:deepcut}
DeepCut aims to solve the problem of multi-person human pose estimation by jointly modeling the poses of all persons appearing in an image. Given an image, it starts by generating a set $D$ of joint proposals, where $\bold{x}_d \in \mathbb{Z}^2$ denotes the 2D location of the $d^{th}$ proposal. The proposals are then used to formulate a graph optimization problem that aims to select a subset of proposals while suppressing the incompatible proposals, label each selected proposal with a joint type $j \in {1 \dots J}$, and associate them to unique individuals. 

The problem can be solved by integer linear programming (ILP), optimizing over the binary variables $x \in \{0,1\}^{D \times J}$, $y \in \{0,1\}^{\binom D2}$, and $z \in \{0,1\}^{\binom D2 \times J^2}$. For every proposal $d$, a set of variables $\{x_{dj}\}_{j = 1 \dots J}$ is defined where $x_{dj} = 1$ indicates that the proposal $d$ is of body joint type $j$. For every pair of proposals $dd'$, the variable $y_{dd'}$ indicates that the proposals $d$ and $d'$ belong to the same person. The variable $z_{dd'jj'} = 1$ indicates that the proposal $d$ is of  joint type $j$, the proposal $d'$ is of joint type $j'$, and both proposals belong to the same person $(y_{dd'} = 1)$. The variable $z_{dd'jj'}$ is constrained such that $z_{dd'jj'} = x_{dj}x_{d'j'}y_{dd'}$. The solution of the ILP problem is obtained by optimizing the following objective function: 
\begin{equation}
\min_{(x,y,z) \in X_{D}} \left\langle \alpha, x \right\rangle + \left\langle \beta, z \right\rangle \label{eq:ilp_obj}
\end{equation}
subject to 
\begin{align}
\forall d \in D~\forall jj' \in \binom J2 :& \quad x_{dj} + x_{dj'} \leq 1 \label{eq:con1}\\
\forall dd' \in \binom D2 :& \quad y_{dd'} \leq \sum_{j \in J} x_{dj}, \quad y_{dd'} \leq \sum_{j \in J} x_{d'j} \label{eq:con2} \\
\forall dd'd'' \in \binom D3 :& \quad y_{dd'} + y_{d'd''} - 1 \leq y_{dd''} \label{eq:con3}\\
\forall dd' \in \binom D2~\forall jj' \in J^2 :& \quad x_{dj} + x_{d'j'} + y_{dd'} - 2 \leq z_{dd'jj'} \nonumber \\
& \quad z_{dd'jj'} \leq min(x_{dj}, x_{d'j'}, y_{dd'}) \label{eq:con4} \\
\shortintertext{and, optionally, } \nonumber \\
\forall dd' \in \binom D2~\forall jj' \in J^2 :& \quad x_{dj} + x_{d'j'} -1 \leq y_{dd'} \label{eq:con5}
\end{align}
where
\begin{align}
\alpha_{dj} &= \log \dfrac{1-p_{dj}}{p_{dj}} \label{eq:unary} \\
\beta_{dd'jj'} &= \log \dfrac{1-p_{dd'jj'}}{p_{dd'jj'}} \label{eq:binary} \\
\left\langle \alpha, x \right\rangle &= \sum_{d \in D} \sum_{j \in J} \alpha_{dj} x_{dj} \\
\left\langle \beta, z \right\rangle &= \sum_{dd' \in \binom D2} \sum_{j,j' \in J} \beta_{dd'jj'} z_{dd'jj'}.
\end{align}
The constraints (\ref{eq:con1})-(\ref{eq:con4}) enforce that optimizing (\ref{eq:ilp_obj}) results in valid body pose configurations for one or more persons. The constraints (\ref{eq:con1}) ensure that a proposal $d$ can be labeled with only one joint type, while the constraints (\ref{eq:con2}) guarantee that any pair of proposals $dd'$ can belong to the same person only if both are not suppressed, \ie, $x_{dj} = 1$ and $x_{d'j'} = 1$. The constraints (\ref{eq:con3}) are transitivity constraints and enforce for any three proposals $dd'd'' \in \binom D3$ that if $d$ and $d'$ belong to the same person, and $d'$ and $d''$ also belong to the same person, then the proposals $d$ and $d''$ must also belong to the same person. The constraints (\ref{eq:con4}) enforce that for any $dd' \in \binom D2$ and $jj' \in J^2$, $z_{dd'jj'} = x_{dj}x_{d'j'}y_{dd'}$. The constraints (\ref{eq:con5}) are only applicable for single-person human pose estimation, as they enforce that two proposals $dd'$ that are not suppressed must be grouped together.  In (\ref{eq:unary}), $p_{dj} \in (0,1)$ are the body joint unaries and correspond to the probability of any proposal $d$ being of joint type $j$.  While in (\ref{eq:binary}), $p_{dd'jj'}$ correspond to the conditional probability that a pair of proposals $dd'$ belongs to the same person, given that $d$ and $d'$ are of joint type $j$ and $j'$, respectively. In \cite{pishchulin2015deepcut} this ILP formulation is referred as \textit{Subset Partitioning and Labelling Problem}, as it partitions the initial pool of proposal candidates to unique individuals, labels each proposal with a joint type $j$, and inherently suppresses the incompatible candidates. 

\subsection{Local Joint-to-Person Association}
\label{sec:local_joint_to_person_association}

In contrast to \cite{pishchulin2015deepcut}, we solve the joint-to-person association problem locally for each person. We also do not label generic proposals as part of the ILP formulation since we use a neural network to obtain detections for each joint as described in Section \ref{sec:cpm}.
We therefore start with a set of joint detections $D_J$, where every detection $d_j$ at location $\bold{x}_{d_j} \in \mathbb{Z}^2$ has a known joint type $j \in {1 \dots J}$. Our model requires only two types of binary random variables $x \in \{0,1\}^{D_J}$ and  $y \in \{0,1\}^{\binom {D_J}{2}}$. Here, $x_{d_j} = 1$ indicates that the detection $d_j$ of part type $j$ is not suppressed, and  $y_{{d_j}{{d'}_{j'}}} = 1$ indicates that the detection $d_j$ of type $j$, and the detection ${d'}_{j'}$ of type $j'$ belong to the same person. The objective function for local joint-to-person association takes the form: \begin{equation}
\min_{(x,y) \in X_{D_J}} \left\langle \alpha, x \right\rangle + \left\langle \beta, y \right\rangle \label{eq:silp_obj}
\end{equation}
subject to
\begin{align}
\forall {d_j}{{d'}_{j'}} \in \binom {D_J}{2} :& \quad y_{{d_j}{{d'}_{j'}}} \leq x_{d_j}, \quad y_{{d_j}{{d'}_{j'}}} \leq x_{d'_{j'}} \label{eq:scon1} \\
\forall {d_j}{d'_{j'}}{d''_{j''}} \in \binom {D_J}{3} :& \quad y_{{d_j}{d'_{j'}}} + y_{{d'_{j'}}{d''_{j''}}} - 1 \leq y_{{d_j}{{d''}_{j''}}} \label{eq:scon2}\\
\forall {d_j}{d'_{j'}} \in \binom {D_J}{2} :& \quad x_{d_j} + x_{d'_{j'}} -1 \leq y_{{d_j}{{d'}_{j'}}} \label{eq:scon3}
\end{align}
where 
\begin{align}
\alpha_{d_j} &= \log \dfrac{1-p_{d_j}}{p_{d_j}} \label{eq:sunary} \\
\beta_{{d_j}{d'_{j'}}} &= \log \dfrac{1-p_{{d_j}{d'_{j'}}}}{p_{{d_j}{d'_{j'}}}} \label{eq:sbinary} \\
\left\langle \alpha, x \right\rangle &= \sum_{d_j \in D_J} \alpha_{d_j} x_{d_j} \\
\left\langle \beta, y \right\rangle &= \sum_{{d_j}{d'_{j'}} \in \binom {D_J}{2}} \beta_{{d_j}{d'_{j'}}} y_{{d_j}{d'_{j'}}}.
\end{align}
The constraints (\ref{eq:scon1}) enforce that detection $d_j$ and ${d'_{j'}}$ are connected $(y_{{d_j}{{d'}_{j'}}} = 1)$ only if both are not suppressed, \ie, $x_{d_j} = 1$ and $x_{d'_{j'}} = 1$. The constraints (\ref{eq:scon2}) are transitivity constraints as before and the constraints (\ref{eq:scon3}) guarantee that all detections that are not suppressed belong to the primary person. We can see from (\ref{eq:ilp_obj})-(\ref{eq:con5}) and (\ref{eq:silp_obj})-(\ref{eq:scon3}), that the number of variables are reduced from $({D \times J}+{\binom D2}+{\binom D2 \times J^2})$ to $({D_J}+{\binom {D_J}{2}})$. Similary, the number of constraints is also drastically reduced.  

In (\ref{eq:sunary}), $p_{d_j} \in (0,1)$ is the confidence of the joint detection $d_j$ as probability. We obtain this directly from the score maps inferred by the CPM as $p_{d_j} = \mathrm{f}_{\tau}(s^j_T(\bold{x}_{d_j}))$, where 
			
\begin{equation}
\mathrm{f}_{\tau}(s) = \begin{cases}
         s & \text{if} \quad s \geq \tau \\
         0 & \text{otherwise,}
        \end{cases}
\label{eq:unary_thresh}
\end{equation}
and $\tau$ is a threshold that suppresses detections with a low confidence score. 

In (\ref{eq:sbinary}), $p_{{d_j}{d'_{j'}}} \in (0,1)$ corresponds to the conditional probability that the detection $d_j$ of joint type $j$ and the detection $d'_{j'}$ of joint type $j'$ belong to the same person. For $ j = j' $, it is the probability that both detections $d_j$ and $d'_{j'}$ belong to the same body joint. For $j \neq j'$, it measures the compatibility between two detection candidates of different joint types. Similar to \cite{pishchulin2015deepcut}, we obtain these probabilities by learning discriminative models based on appearance and spatial features of the detection candidates. For $j = j'$, we define a feature vector 
\begin{equation}
f_{{d_j}{d'_{j'}}} = \{\bigtriangleup \bold{x}, \exp (\bigtriangleup \bold{x}), (\bigtriangleup \bold{x})^2\},
\end{equation}  
where $\bigtriangleup \bold{x} = (\bigtriangleup u, \bigtriangleup v)$ is the 2D offset between the locations $\bold{x}_{d_j}$ and $\bold{x}_{d'_{j'}}$.
For $j \neq j'$, we define a separate feature vector based on the spatial locations as well as the appearance features obtained from the joint detectors as 
\begin{equation}
f_{{d_j}{d'_{j'}}} = \{\bigtriangleup \bold{x}, \lVert\bigtriangleup \bold{x}\rVert, \arctan\left(\dfrac{\bigtriangleup v }{\bigtriangleup u}\right), \bold{s}_{T}(\bold{x}_{d_j}), \bold{s}_{T}(\bold{x}_{{d'}_{j'}}) \},
\end{equation} 
where $\bold{s}_{T}(\bold{x})$ is a vector containing the confidences of all joints and the background at location $\bold{x}$.  
For both cases, we gather positive and negative samples from the annotated poses in the training data and train an SVM with RBF kernel using LibSVM \cite{libsvm} for each pair $jj' \in \binom J2$. In order to obtain the probabilities $p_{{d_j}{d'_{j'}}} \in (0,1)$ we use Platt scaling \cite{Platt99probabilisticoutputs} to normalize the output of the SVMs to probabilities. After optimizing \eqref{eq:silp_obj}, the pose of the primary person is given by the detections with $x_{d_j}=1$. 

\section{Experiments}
We evaluate the proposed approach on the Multi-Person subset of the MPII Human Pose Dataset \cite{andriluka_cvpr2014} and follow the evaluation protocol proposed in \cite{pishchulin2015deepcut}. The dataset consists of 3844 training and 1758 testing images with multiple persons. The persons appear in highly articulated poses with a large amount of occlusions and truncations. Since the original test data of the dataset is withheld, we perform all intermediate experiments on a validation set of 1200 images. The validation set is sampled according to the split proposed in \cite{tompson_cvpr2015} for the single person setup, \ie, we chose all multi-person images that are part of the validation test set proposed in \cite{tompson_cvpr2015} and use all other images for training. In addition we compare the proposed method with the state-of-the-art approach \cite{pishchulin2015deepcut} on their selected subset of 288 images, and also compare with \cite{insafutdinov2016deepercut} on the complete test set. The accuracy is measured by average precision (AP) for each joint using the scripts provided by \cite{pishchulin2015deepcut}. 

\subsection{Implementation Details}
In order to localize the persons, we use the person detector proposed in \cite{ren2015faster}. The detector is trained on the Pascal VOC dataset \cite{Everingham15}. For the quantitative evaluation, we discard detected persons with a 
bounding box area less equal to $80\times80$ pixels since small persons are not annotated in the MPII Human Pose Dataset. For the qualitative results shown in Figure \ref{fig:qualitative_results}, we do not discard the small detections. 
For the CPM \cite{wei2016convolutional}, we use the publicly available source code and train it on the Multi-Person subset of the MPII Human Pose Dataset as described in Section \ref{sec:cpm}. As in \cite{wei2016convolutional}, we add images from the Leeds Sports Dataset \cite{Ever10} during training, and use a 6 stage $(T=6)$ CPM architecture. For solving \eqref{eq:silp_obj}, we use the Gurobi Optimizer.

\subsection{Results}

\begin{figure*}[t!]
\centering
\captionsetup[figure]{skip=0pt}
\includegraphics[scale=0.8]{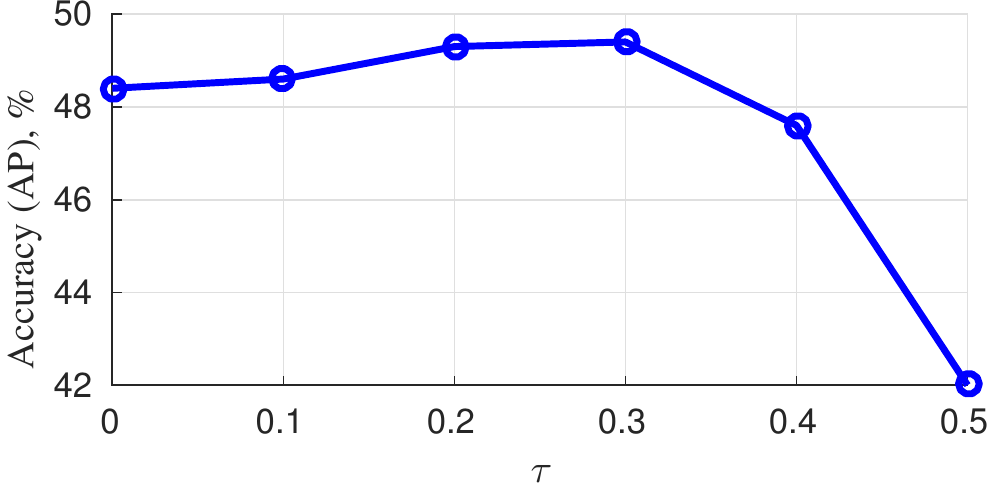}
\caption{Impact of the parameter $\tau$ in (\ref{eq:unary_thresh}) on the pose estimation accuracy.}
\label{fig:tau_plot}
\end{figure*}

We first evaluate the impact of the parameter $\tau$ in $f_{\tau}(s)$ \eqref{eq:unary_thresh} on the pose estimation accuracy measured as mean AP on the validation set containing 1200 images. Figure \ref{fig:tau_plot} shows that the function $f_{\tau}$ improves the accuracy when $\tau$ is increased until $\tau = 0.3$. For $\tau > 0.4$, the accuracy drops since a high value discards correct detections. For the following experiments, we use $\tau = 0.2$.

\begin{table*}[t]
  \centering
  \footnotesize
  \setlength{\tabcolsep}{2.8pt} 
\def\arraystretch{1}

\scalebox{0.95}{

\begin{tabular}{*{9}{lcccccccc|c}}
\hline
Setting  &Head & Shoulder & Elbow & Wrist & Hip & Knee  & Ankle & Total & time (s)\\ \hline

CPM only & 56.5  & 55.2  & 47.6  & 39.1  & 48.1  & 39.6 & 30.3 & 45.2 &  2\\

CPM+L-JPA (N=1) & 59.9  & 57.5  & 50.5  & 41.8  & 49.8  & 45.5 & 39.5 & 49.2 & 3 \\

CPM+L-JPA (N=3) & 59.9  & 57.4  & 50.7  & 42.1  & 50.0  & 45.5 & 39.5 & 49.3 & 8\\

CPM+L-JPA (N=5) & 60.0  & 57.5  & 50.7  & 42.1  & 50.1  & 45.5 & 39.4 & 49.3 & 10 \\  \hline

CPM+L-JPA (N=5)+GT Torso & 92.9  & 91.3  & 78.4  & 61.8  & 81.0  & 71.4 & 61.8 & 76.9 & 10\\

	\hline
  \end{tabular}
  }
  \caption{Pose estimation results (AP) on the validation test set (1200 images) of the MPII Multi-Person Pose Dataset.}
\label{tab:final_acc_mpi_val}
\end{table*}

Table \ref{tab:final_acc_mpi_val} reports the pose estimation results under different settings of the proposed approach on the validation set. We also report the median run-time required by each setting\footnote{Measured on a 2GHz Intel(R) Xeon(R) CPU with a single core and NVidia Geforce GTX Titan-X GPU.}. Using only the CPM to estimate the pose of each detected person achieves 45.2\% mAP and takes only 2 seconds per image. Using the proposed Local Joint-to-Person Association (L-JPA) model with 1 detection candidate per joint $(N=1)$ to suppress the incompatible detections improves the performance from $45.2\%$ to $49.2\%$ with a very slight increase in run-time. Increasing the number of candidates per joint increases the accuracy only slightly. For the following experiments, we use $N=5$. When we compare the numbers with Figure \ref{fig:tau_plot}, we observe that CPM+L-JPA outperforms CPM for any $0 \leq \tau \leq 0.4$.

The accuracy also depends on the used person detector. We use an off-the-shelf person detector without any fine-tuning on the MPII dataset. In order to evaluate the impact of the person detector accuracy, we also estimate poses when the person detections are given by the ground-truth torso (GT Torso) locations of the persons provided with the dataset. 
This results in a significant improvement in accuracy from $49.3\%$ to $76.9\%$ mAP, showing that a better person detector would improve the results further.

\begin{table*}[t]
  \centering
  \footnotesize
  \setlength{\tabcolsep}{2.8pt} 
\def\arraystretch{1}

\scalebox{1}{

  \begin{tabular}{*{9}{lcccccccc|c}}
\hline
Setting  &Head & Shoulder & Elbow & Wrist & Hip & Knee  & Ankle & Total & time (s)\\ \hline
Ours & 70.0  & 65.2  & 56.4  & 46.1  & 52.7 & 47.9 & 44.5 & 54.7 & 10 \\ 
DeepCut \cite{pishchulin2015deepcut} & 73.1  & 71.7  & 58.0  & 39.9  & 56.1  & 43.5 & 31.9 & 53.5 & 57995  \\
DeeperCut \cite{insafutdinov2016deepercut} & 87.9  & 84.0  & 71.9 & 63.9  & 68.8  & 63.8 & 58.1 & 71.2 & 230 \\
	\hline
Ours GT ROI & 87.7  & 81.6  & 68.9  & 56.1  & 66.4  & 59.4 & 54.0 & 67.7 & 10 \\ 
DeepCut GT ROI \cite{pishchulin2015deepcut} & 78.1  & 74.1  & 62.2  & 52.0  & 56.9  & 48.7 & 46.1 & 60.2 & -  \\
Chen et al., GT ROI \cite{chen_nips2014} & 65.0& 34.2 & 22.0 & 15.7 & 19.2 & 15.8 & 14.2 & 27.1 & -	\\ \hline
  \end{tabular}
  }
  \caption{Comparison of pose estimation results (AP) with state-of-the-art approaches on 288 images \cite{pishchulin2015deepcut}.}
  \label{tab:final_acc_288_images}
\end{table*}

Table \ref{tab:final_acc_288_images} compares the proposed approach with other approaches on a selected subset of 288 test images used in \cite{pishchulin2015deepcut}. Our approach outperforms the state-of-the-art method DeepCut \cite{pishchulin2015deepcut} ($54.7\%$ vs.~$53.5\%$) while being significantly faster (10 seconds vs.~57995 seconds). If we use $N=1$, our approach requires only 3 seconds per image with a minimal loss of accuracy as shown in Table \ref{tab:final_acc_mpi_val}, \ie, our approach is more than 19,000 times faster than \cite{pishchulin2015deepcut}. We also compare with a concurrent work \cite{insafutdinov2016deepercut}. While the approach \cite{insafutdinov2016deepercut} achieves a higher accuracy than our method, our method is significantly faster (10 seconds vs.~230 seconds). In contrast to \cite{insafutdinov2016deepercut}, we do not perform fine-tuning of the person detector on the MPII Multi-Person Pose Dataset and envision that doing this will lead to further improvements. We therefore compare with two additional approaches \cite{pishchulin2015deepcut, chen_nips2014} when using GT bounding boxes of the persons. The results for \cite{chen_nips2014} are taken from \cite{pishchulin2015deepcut}. Our approach outperforms both methods by a large margin.

\begin{table*}[t]
  \centering
  \footnotesize
  \setlength{\tabcolsep}{2.8pt} 
\def\arraystretch{1}

\scalebox{1}{
  \begin{tabular}{*{9}{lcccccccc|c}}
\hline
Setting  &Head & Shoulder & Elbow & Wrist & Hip & Knee  & Ankle & Total & time (s)\\ \hline
Ours & 58.4  & 53.9  & 44.5  & 35.0  & 42.2 & 36.7 & 31.1 & 43.1 & 10 \\
DeeperCut \cite{insafutdinov2016deepercut} & 78.4  & 72.5  & 60.2 & 51.0  & 57.2  & 52.0 & 45.4 & 59.5 & 485\\ 
	\hline
	\multicolumn{10}{c}{Using GT ROIs} \\
	\hline
Ours + GT Torso & 85.6  & 79.4  & 62.9  & 48.9  & 62.6  & 51.9 & 43.9 & 62.2 & 10 \\ 
	\hline
  \end{tabular}
  }
  \caption{Pose estimation results (AP) on the withheld test set of the MPII Multi-Person Pose Dataset.}
  \label{tab:final_acc_full_mpi}
\end{table*}

\begin{figure*}
  \centering
  \begin{tabular}{c c c c}
    \includegraphics[height=0.170\linewidth]{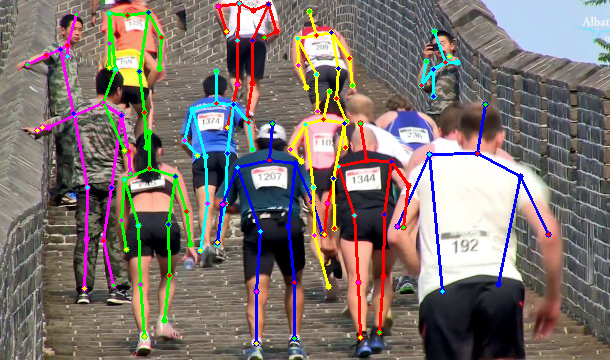} &
    \includegraphics[height=0.170\linewidth]{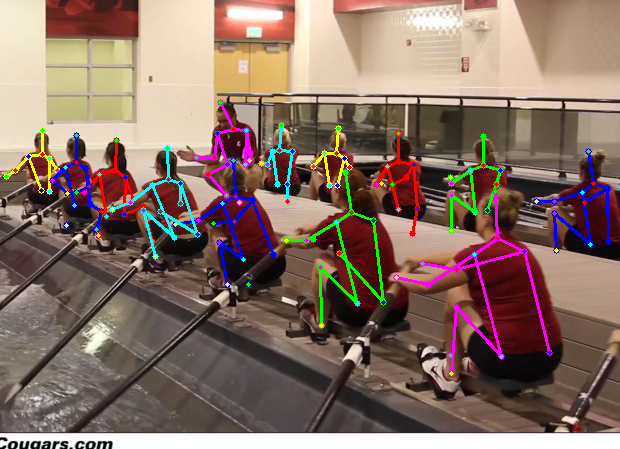} &
    \includegraphics[height=0.170\linewidth]{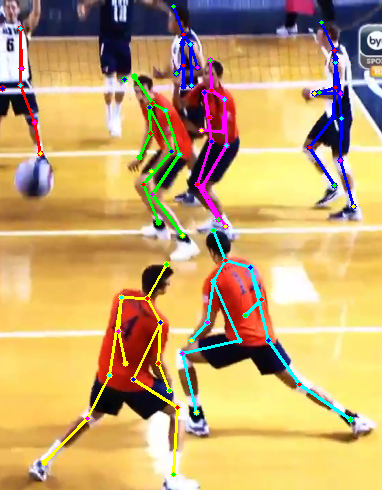} &
    \includegraphics[height=0.170\linewidth]{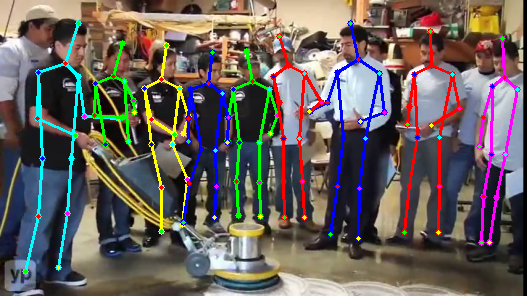} \\
  \end{tabular}

  \begin{tabular}{c c c c}
    \includegraphics[height=0.164\linewidth]{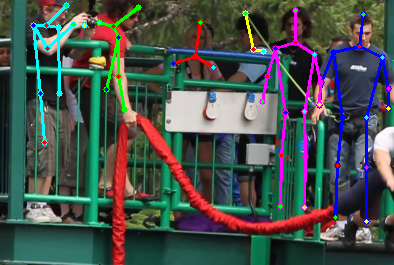} &
    \includegraphics[height=0.164\linewidth]{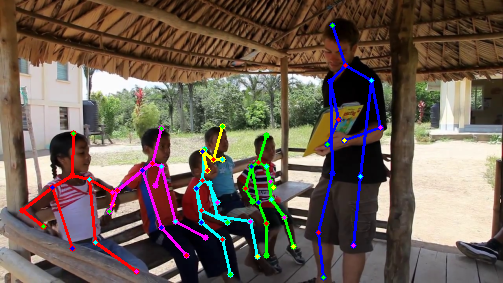} &
    \includegraphics[height=0.164\linewidth]{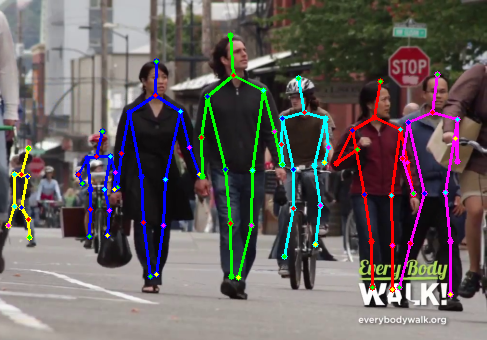} &
    \includegraphics[height=0.164\linewidth]{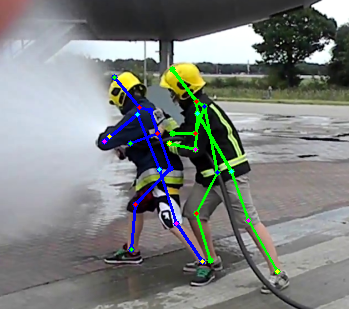} \\
  \end{tabular}
  \begin{tabular}{c c c c}
    \includegraphics[height=0.179\linewidth]{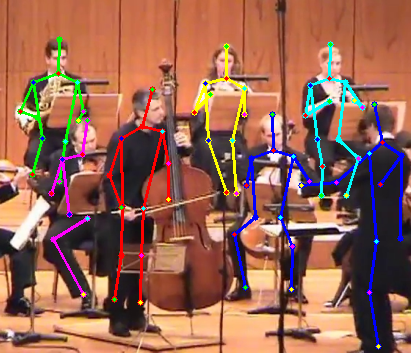} &
    \includegraphics[height=0.179\linewidth]{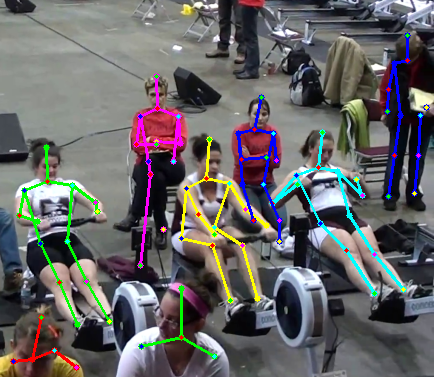} &
    \includegraphics[height=0.179\linewidth]{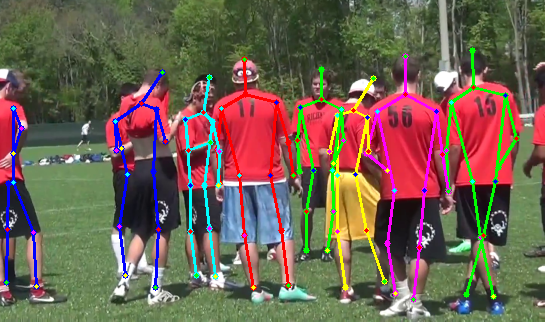} &
    \includegraphics[height=0.179\linewidth]{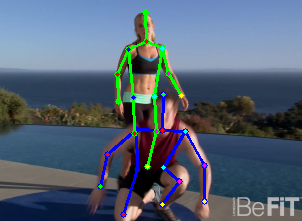}\\
  \end{tabular}

  \begin{tabular}{c c c c}
    \includegraphics[height=0.181\linewidth]{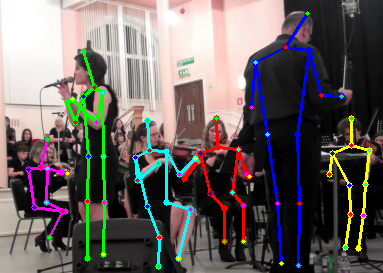} &
    \includegraphics[height=0.181\linewidth]{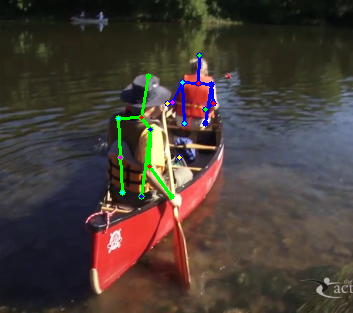} &
    \includegraphics[height=0.181\linewidth]{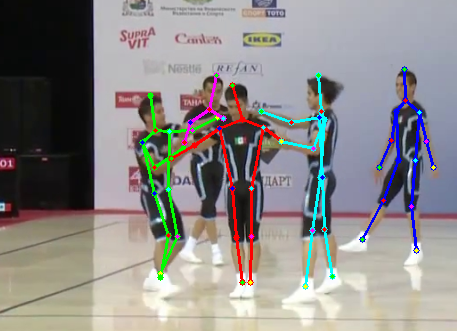} &
    \includegraphics[height=0.181\linewidth]{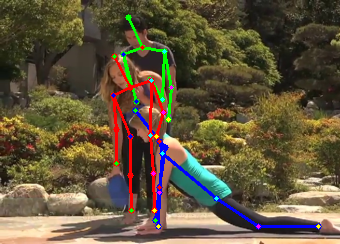} \\
  \end{tabular}

  \begin{tabular}{c c c c}
    \includegraphics[height=0.180\linewidth]{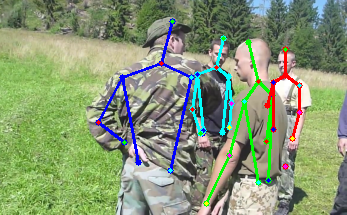} &
    \includegraphics[height=0.180\linewidth]{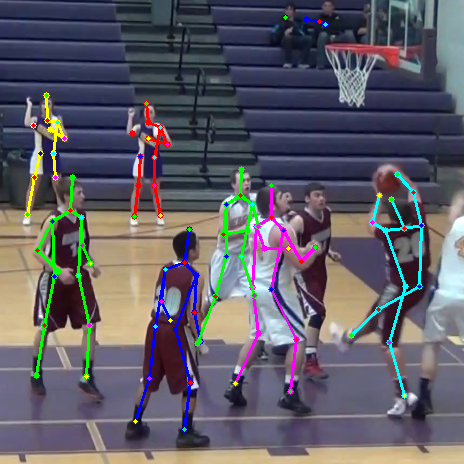} &
    \includegraphics[height=0.180\linewidth]{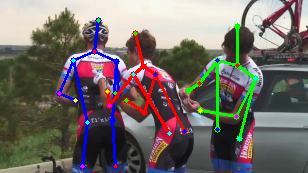} & 
    \includegraphics[height=0.180\linewidth]{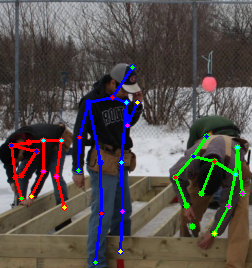} \\
  \end{tabular}

  \begin{tabular}{c c c c}
    \includegraphics[height=0.173\linewidth]{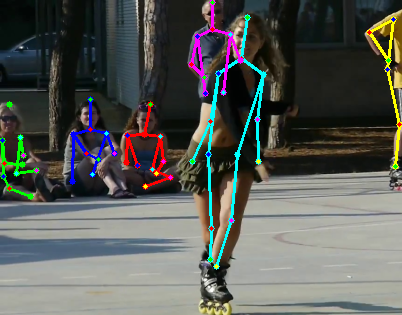} &
    \includegraphics[height=0.173\linewidth]{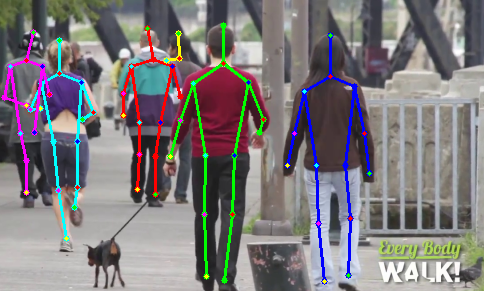} &
    \includegraphics[height=0.173\linewidth]{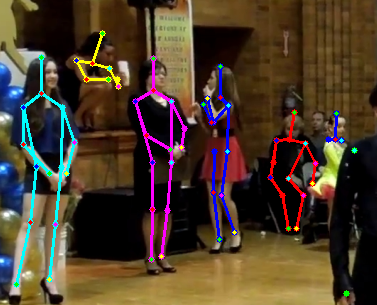} &
    \includegraphics[height=0.173\linewidth]{0780} \\
  \end{tabular}

  \begin{tabular}{c c c c}
    \includegraphics[height=0.162\linewidth]{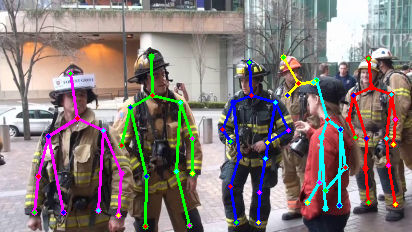} &
    \includegraphics[height=0.162\linewidth]{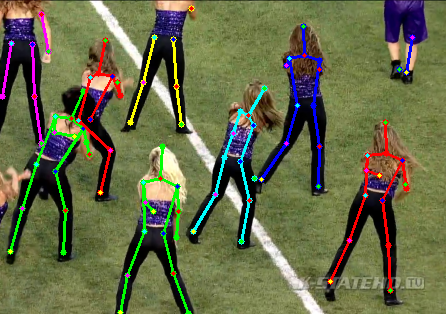} &
    \includegraphics[height=0.162\linewidth]{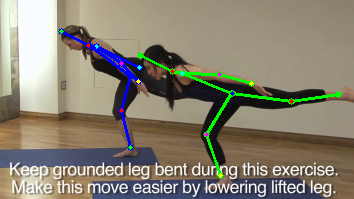} &
    \includegraphics[height=0.162\linewidth]{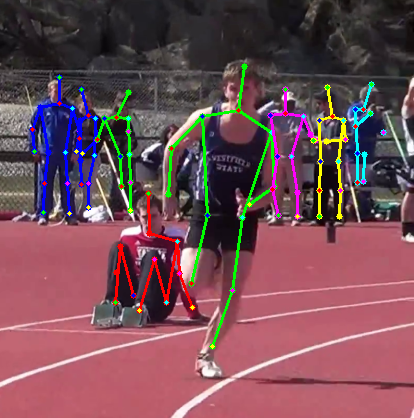} \\
  \end{tabular}
  \vspace{-1em}
  \caption{Some qualitative results for the MPII Multi-Person Pose Dataset.}
  \label{fig:qualitative_results}
\end{figure*}

Finally in Table \ref{tab:final_acc_full_mpi}, we report our results on all test images of the MPII Multi-Person Pose Dataset. Our method achieves $43.1\%$ mAP. While the approach \cite{pishchulin2015deepcut} cannot be evaluated on all test images due to the high computational complexity of the model, \cite{insafutdinov2016deepercut} reports a higher accuracy than our model. However, if we compare the run-times in Table \ref{tab:final_acc_288_images} and Table \ref{tab:final_acc_full_mpi}, we observe that the run-time of \cite{insafutdinov2016deepercut} doubles on the more challenging test set (485 seconds per image). Our approach on the other hand requires only 10 seconds in all evaluation settings and is around 50 times faster. If we use $N=1$, our approach is 160 times faster than \cite{insafutdinov2016deepercut}. Using the torso annotation (GT Torso) as person detections results again in a significant improvement of the accuracy ($62.2\%$ vs.~$43.1\%$ mAP). Some qualitative results can be seen in Figure \ref{fig:qualitative_results}.

\section{Conclusion}
In this work we have presented an approach for multi-person pose estimation under occlusions and truncations. Since the global modeling of poses for all persons is impractical, we demonstrated that the problem can be formulated by a set of independent local joint-to-person association problems. Compared to global modeling, these problems can be solved efficiently while still being effective for handling severe occlusions or truncations. Although the accuracy can be further improved by using a better person detector, the proposed method already achieves the accuracy of a state-of-the-art method, while being 6,000 to 19,000 times faster. 

\subsubsection*{Acknowledgements:}
The work was partially supported by the ERC Starting Grant ARCA (677650).

\bibliographystyle{splncs}
\bibliography{pose_bib}
\end{document}